\documentclass[journal,twoside]{IEEEtran}
\usepackage{cite}
\usepackage{amsmath,amssymb,amsfonts, dsfont}
\usepackage{algorithmic}
\usepackage{algorithm}
\usepackage{graphicx}
\usepackage{hyperref}
\usepackage{textcomp}
\usepackage{multirow}
\usepackage{adjustbox}

\begin{document}
\title{Surg - InvNeRF: Invertible NeRF for 3D tracking and reconstruction in surgical vision}
\author{Gerardo Loza, Junlei Hu, Dominic Jones, Sharib Ali and Pietro Valdastri 
\thanks{This work was supported by the National Institutes of Health (NIH) (Award No. R01EB018992). Any opinions, findings, conclusions, or recommendations expressed in this material are those of the authors and do not necessarily reflect the views of the NIH. The work was also supported in part by UK Research and Innovation (UKRI) [CDT grant number EP/S024336/1] and by the Engineering and Physical Sciences Research Council [grant number UKRI914]. }
\thanks{G. Loza's Ph.D. studies is supported by the Mexican Council for Science and Technology (CONACYT) and the University of Leeds. This work was undertaken on the Aire HPC system at the University of Leeds, UK.}
\thanks{G. Loza and S. Ali are with the School of Computer Science, Faculty of Engineering and Physical Sciences, University of Leeds, LS2 9JT, West Yorkshire, UK. J. Hu, D. Jones and P. Valdastri are with the School of Electronic and Electrical Engineering, Faculty of Engineering and Physical Sciences, University of Leeds, LS2 9JT, West Yorkshire, UK; email: \{scgelg, eljh, d.p.jones, s.s.ali, p.valdastri\}@leeds.ac.uk} 
\thanks{P. Valdastri and S. Ali: Shared senior authorship 
}
}

\maketitle
\begin{abstract}
We proposed a novel test-time optimisation (TTO) approach framed by a NeRF-based architecture for long-term 3D point tracking. Most current methods in point tracking struggle to obtain consistent motion or are limited to 2D motion. TTO approaches frame the solution for long-term tracking as optimising a function that aggregates correspondences from other specialised state-of-the-art methods. Unlike the state-of-the-art on TTO, we propose parametrising such a function with our new invertible Neural Radiance Field (InvNeRF) architecture to perform both 2D and 3D tracking in surgical scenarios. 
Our approach allows us to exploit the advantages of a rendering-based approach by supervising the reprojection of pixel correspondences.
It adapts strategies from recent rendering-based methods to obtain a bidirectional deformable-canonical mapping, to efficiently handle a defined workspace, and to guide the rays' density. It also presents our multi-scale HexPlanes for fast inference and a new algorithm for efficient pixel sampling and convergence criteria.  We present results in the STIR and SCARE datasets, for evaluating point tracking and testing the integration of kinematic data in our pipeline, respectively. In 2D point tracking, our approach surpasses the precision and accuracy of the TTO state-of-the-art methods by nearly 50\% on average precision, while competing with other approaches. In 3D point tracking, this is the first TTO approach, surpassing feed-forward methods while incorporating the benefits of a deformable NeRF-based reconstruction. 
\end{abstract}


\section{Introduction}
\label{sec:introduction}
\IEEEPARstart{T}{racking} points and surfaces in 3D for robotic minimally invasive surgery (RMIS) is essential for developing solutions to various downstream applications~\cite{13_Bcm,10_Br}, including force estimation, skill assessment, augmented reality, guided navigation, and autonomy. 
Solutions for these tasks must address frequent surgical-related problems, such as intermittent blurriness, variable illumination, complex tissue deformations, sight loss, and homogeneous texture~\cite{12_Bcm}. 

Considerable effort has been made to break down these tasks into more elemental steps, establishing a general workflow~\cite{14_Bc}: image formation, feature detection, feature description, point matching, and finally point tracking (PT) and depth estimation (DE). In this way, research has focused on improving one or multiple stages of this workflow~\cite{16_Bcm,04_Wt,05_Wt, 45_Wt,49_Wt}. 
Deep learning methods have made significant progress on these tasks to improve tracking of relevant tissues by relying on ground-truth 3D information, stereo images, sequential frames, or camera pose to supervise the learning~\cite{36_Wd, 40_Wd, 41_Wd, 46_Wt}. However, the reliability of these approaches is often strongly constrained to short image sequences. 
On top, many approaches have incorporated the camera location as part of the problem to address simultaneously tracking and reconstruction~\cite{50_Wd}. Ideas such as \textit{Structure from motion} (SfM) and \textit{Simultaneous Localisation and Mapping} (SLAM), where the position of the camera for the reconstruction is estimated, have become popular~\cite{53_M,54_Wdt,55_Wdt}. In general, the inclusion of camera pose estimation has been shown to provide additional 3D context~\cite{35_Wd, 37_Wd, 43_Wd, 07_Wd}. 

Recent rendering-based methods, such as NeRF~\cite{70_M} and Gaussian Splatting~\cite{21_M}, also rely on camera poses and obtain a reconstruction of the 3D space, without explicitly matching pixels from different images. Most current rendering-based methods in surgery focus on novel view synthesis~\cite{06_Wd,71_M,72_W}, despite the fact that their implicit strong point-matching stage, and the capability to model deformations, can be exploited to have a more complete scene representation that 3D tracks the movements of the objects. 
Outside the surgical domain, Omnimotion~\cite{17_M} is an approach inspired by NeRF that sacrifices the implicit 3D reconstruction of the space by assuming an orthographic camera model and ignoring the relationship between the rays' density and the 3D object position to build a framework capable of test-time optimisation for global 2D motion (long-term 2D tracking from optical flow aggregation). However, it has not been used for tissue tracking.

In this context, the assumption of simple and smooth deformations, which demonstrated promising results in short image sequences, does not hold up in longer sequences. Furthermore, most current approaches are limited to 2D tracking and do not incorporate any form of correspondence aggregation to enhance results on long sequences. Therefore, new test-time optimisation approaches that leverage the aggregation of 2D motion and accurately extrapolate to the 3D space to 3D track deformable objects, while allowing for the incorporation of camera poses from a robotic system to disentangle the camera movement from the tissue deformation, are yet to be developed to adequately address the challenges of long-term 3D tracking in a surgical environment. Our contribution is an invertible rendering-based framework called Invertible Neural Radiance Field (InvNeRF) for test-time optimisation (TTO) in 3D. The key contributions of our work are summarised below:

\begin{itemize}
    \item We propose a novel TTO approach for long-term 3D point tracking by aggregating stereo depth estimation and reliable 2D short-term correspondences.
    \item We propose an efficient pixel sampling algorithm to handle redundant sampling and enable faster optimisation.
    \item A novel joint loss function is introduced to exploit sub-spaces, including image 2D space, deformable 3D space (workspace) and canonical 3D space, for projection and re-projection error minimisation. 
    \item We explore integration of semantic information (such as tool masking), and disentanglement of camera motion and tissue deformation by informing camera pose through kinematics data. These processes simplify optimisation by focusing on the target tissue only.
    \item We validate our approach for both 2D and 3D point tracking using publicly available STIR \cite{01_Dt} and SCARED~\cite{69_Dd} datasets. Experimental results demonstrate our approach is comparable to the SOTA approaches for 2D and significantly outperforms 3D point tracking.
\end{itemize}

\section{Related work}
This section describes recent approaches for consistent pixel correspondences on long surgical image sequences. 
\subsection{Feed-forward approaches}
An unsupervised deep tracking model using a bidirectional Siamese neural network was introduced to address the lack of labels in the datasets \cite{04_Wt}. However, this approach struggles when tissues are occluded or undergo complex deformations. SENDD~\cite{03_Wt} is one of the recent feed-forward deep-learning approaches applied to tissue in a surgical environment that focuses on both tracking and stereo depth estimation. It integrates graph neural networks (GNN) to detect and describe sparse, reliable correspondences. It reports an average 3D tracking error of 7.92 $mm$, but does not provide code to reproduce the results or a comparison. Additionally, it has limitations when handling occlusions and drift over long image sequences~\cite{03_Wt}.
Endo-Depth-and-Motion~\cite{07_Wd} presents a method for camera pose estimation and reconstruction with a self-supervised framework. It uses Monodepth2~\cite{monodepth2}, photometric tracking, and volumetric fusion to obtain depth information, camera position, and reconstruction, respectively, under the assumption of smooth and slow deformations. A SLAM-based method and the modelling of physiological motion for compensating the patient's respiration during the point-matching stage was presented in \cite{63_Wdt}; however, it assumes rigid registration.
A dataset for validating SLAM methods on endoscopic images was introduced in \cite{08_Wd}. The authors also  proposed a self-supervised approach ``Endo-SfMLearner'', which is a combination of residual networks and attention mechanisms to predict camera pose and depth information. Similarly, other works have addressed the problem by incorporating camera pose estimation methods into depth estimators for single frames, stereo frames, or pairs of consecutive mono frames~\cite{60W_dt,61_Wdt,62_W_dt,64_Wdt}. 
These methods have made outstanding progress by diving into one of the next steps of surgical understanding, 3D reconstruction; however, the challenges and flaws from the point tracking (PT) and depth estimation  (DE) stages can hinder their performance. Problems include illumination, blurriness and anisotropic tissue deformations that adversely affects features estimation over long sequence; similarly, occlusion and fast movements contradict the assumption of continuity, smoothness and slow motion that many of these methods use; and finally, the disentanglement of camera motion from tissue deformations is a complex task that current methods have not been able to address. 

Recent works have recognised the short-term consistency limit in most current DE and PT methods and proposed some alternatives. \cite{65_Wd} proposed a self-supervised framework to enhance the temporal consistency of depth models, by guiding training with stereo depth and image warping. Their results demonstrate temporally consistent depth estimation, but the final output is scale-agnostic (i.e., no metric distances are predicted by the model), so the information can not be directly used for 3D tracking after finding 2D correspondences. 

\subsection{Render-based approaches}
D-NeRF for deformable surgical scenes~\cite{06_Wd} presents a rendering-based approach for scene reconstruction. They preprocess the depth maps with STTR-light~\cite{67_M} and use them as ground truth during training; they also mask out the tools, providing depth maps of the surgical scenario even through occlusion. They report promising results on novel image synthesis by accurately encoding tissue deformation, but the extraction of correspondences from the implicit representation remains to be explored.
\cite{74_Wt} presents a rendering-based approach for online 3D reconstruction and tracking, by growing and transforming a Gaussian splatting representation. Both transforms and Gaussian representation are optimised per time step (transformations are chained). 2D tracking results are reported with an average error of 11 pixels for six manually selected videos (StereoMIS dataset), but there is no strong validation for the reconstruction or 3D tracking. The reconstruction was not compared against any 3D ground truth information, and there is no description of the criteria used in the pixel correspondence labelling process.
\cite{66_Wt} presents an implementation of Omnimotion~\cite{17_M} with a custom loss function for rigid objects, as their focus was on 2D tracking points on surgical instruments. They reported 74\% average position accuracy, but their assumption about rigid objects can not be applied to tissues on long sequences, and the training times are lengthy.

In general, rendering-based methods offer a new 3D representation of objects with a learnable and data-driven optimisation process, unlike SLAM, which is typically an analytical and deterministic optimisation process. In this work, we introduce a new render-based method for deformable scenarios that exploits data-driven adaptability and can handle the non-linear, fast, and often unpredictable tissue deformations. Furthermore, by supervising implicit pixel correspondences through reliable short-term matches, we enable long-term tracking without violating the assumptions of rigid objects and smooth motion within short time windows, conditions under which the tissue is most likely to exhibit such behaviour.
Finally, we introduce rendering-based approach that directly incorporates the kinematic data from a robotic system that can explicitly disentangle camera movements from tissue deformations.

\subsection{Problem definition} 
We define the problem of establishing frame-to-frame correspondences for a long sequence of consecutive video frames as follows. 
Let $\mathrm{\textbf{I}}=\{I_0,I_1,I_2,...,I_t\}$ denote the sequence of RGB images, 
where $I_t\in\mathds{R}^{H,W,3}$ for every time $t$. Let $\textbf{p}_t$ represent pixel locations within $I_t$. The function $f_{ST}$ can be any feature matcher function that identifies reliable short-term correspondences between frames, predicting the location $\textbf{p}_{t_1 \rightarrow t_2}$ in $I_{t_2}$ matching $\textbf{p}_{t_1}$ in $I_{t_1}$, as defined by $\textbf{p}_{t_1 \rightarrow t_2} = f_{ST}(\textbf{p}_{t_1}, I_{t_1}, I_{t_2})$ such that $||t_2-t_1||<k$ where $k$ is the maximum time window between short-term pairs.
Given a set $\Omega$ of short-term correspondences $(\textbf{p}_{t_1}, \textbf{p}_{t \rightarrow t_2})$, we look for a long-term correspondence function $f_{LT}(\textbf{p}_{t_1},t_1,t_2)=\hat{\textbf{p}}_{t_1 \rightarrow t_2}$ that minimise the error between $\hat{\textbf{p}}_{t_1 \rightarrow t_2}$ and multiple chains to go from $t_1$ to $t_2$ using elements of $\Omega$.
 
Omnimotion~\cite{17_M} proposed a rendering-based framework for the parametrisation of $f_{LT}$. However, this approach assumes an orthographic camera model and an unbounded temporal window for short-term pairs making it computationally expensive and inefficient. Thus, the current SOTA does not provide reliable correspondences from $f_{ST}$ when $|t_2|>>|t_1|$ and the 3D rendered positions do not represent actual locations in the workspace. Along these lines, FastOmniTrack~\cite{77_M} presented an improvement in speed and accuracy by removing the rendering stage, using precomputed depth, reducing the size of the temporal window, and introducing a non-linear invertible function. Despite the improvement, it retains the orthographic camera model and utilises non-metric depth, which makes it impossible to 3D track points in the scene. In this work, we propose the parametrisation of $f_{LT}$ using a pinhole camera model in our Invertible NeRF to fully recover the benefits of a 3D reconstruction and 3D track points. We also reduce memory requirements by constraining the maximum number of pairs for each time step to speed up the optimisation of the function (i.e., reduce the number of iterations).

\section{Methodology}
\begin{figure*}[t!]
\centering
\includegraphics[width=7in]{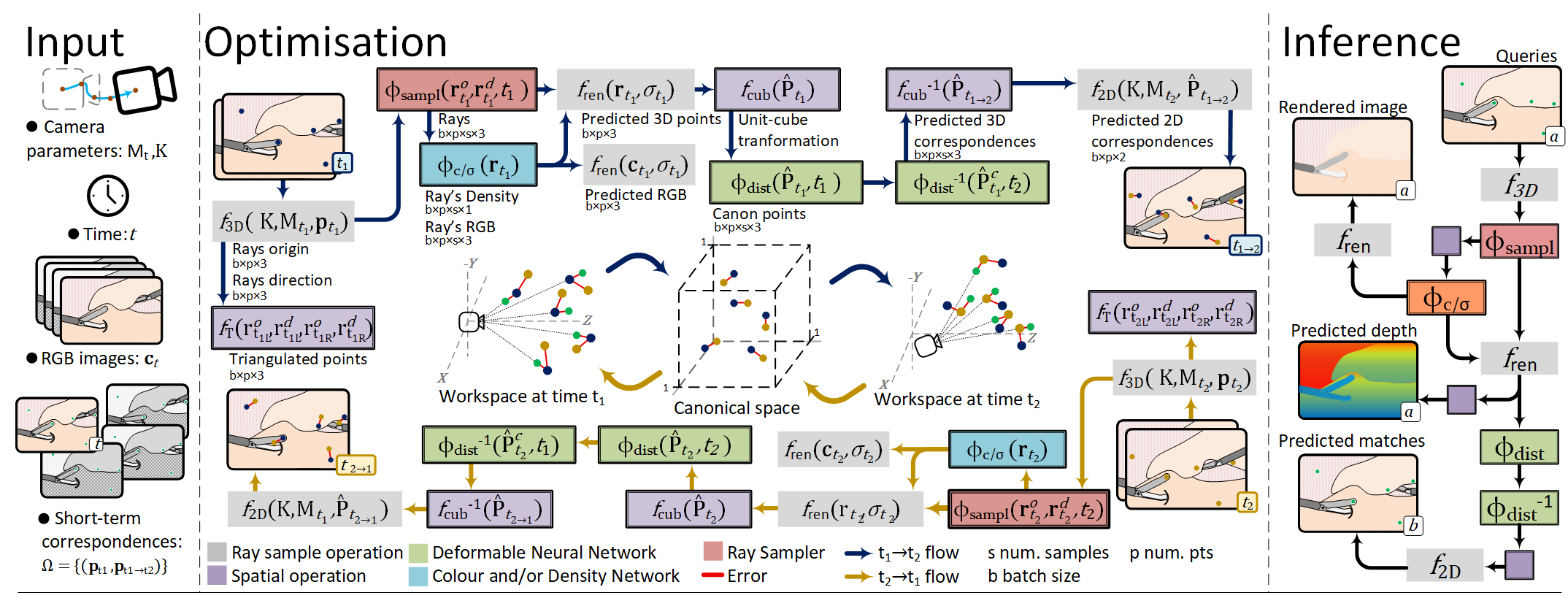}
\caption{
\textbf{Input.} For each given pair of images from different time steps ($t_1 \neq t_2$), the input is the camera parameters and a set of matched pixel coordinates, with their colours and 2D location on the image planes.  
\textbf{Optimisation.} The information from each time-step flows through the networks bidirectionally ($1\rightarrow2, 2\rightarrow1$). Stereo correspondences are used to triangulate 3D points that guide the rays' density (Green points). Rays from the left frames are sampled by $\phi_{sampl}$, $\phi_{c/\sigma}$ predicts colour and density for each sample, and the final colour ($\hat{\textbf{c}}_t$) and 3D location ($\hat{\textbf{P}}_t$) for each pixel are rendered. These 3D points are mapped from a deformable workspace to a static canonical space ($\hat{\textbf{P}}^c_t$), then back to the deformable space at a different time step ($\hat{\textbf{P}}_{t_1 \rightarrow t_2}, \hat{\textbf{P}}_{t_2 \rightarrow t_1}$), and projected to the image plane ($\hat{\textbf{p}}_{t_1 \rightarrow t_2}, \hat{\textbf{p}}_{t_2 \rightarrow t_1}$). Then the error in each space (red lines), image plane, workspace and canonical space is used to optimise the model.
\textbf{Inference.} The optimised NeRF representation holds all the information of the scene, so queries can be sparse points, dense patches, or just camera pose and time to render an image, depth or find pixel correspondences along the whole video.}
\label{fig:1}
\end{figure*}

This section describes our proposed NeRF-based framework for test-time optimisation that aggregates the 3D projection of short-term 2D pixel correspondences for long-term 3D tracking. An overview of our proposed framework's input, optimisation and inference flow is shown in Fig.~\ref{fig:1}. Here,$f_{LT}$ is mainly defined by the combination of three core components: 1) a ray sampler $\phi_{sampl}$ (red block), 2) a colour and density network ${\phi}_{c/\sigma}$ (blue block), and 3) an invertible deforming network $\phi_{def}$ (green block). 
During the optimisation, the data from pairs of frames ($I_{t_1}$ and $I_{t_2}$) with short-term pixel correspondences in $\Omega$ is passed. It includes frames' time steps $(t_1, t_2)$, 2D pixel locations $(\textbf{p}_{t_1}, \textbf{p}_{t_2})$, pixel colours $(\textbf{c}_{t_1},\textbf{c}_{t_2})$, extrinsic camera parameters $(\mathrm{M}_{t_1}, \mathrm{M}_{t_2})$ and the intrinsic camera parameters $(\mathrm{K})$.Additionally, the stereo correspondences of $\textbf{p}_{t_{1R}}$ and $\textbf{p}_{t_{2R}}$ are passed. Rays from all points are calculated by $f_{3\mathrm{D}}(p_t, \mathrm{M}_{t}, \mathrm{K}) = \mathrm{\textbf{r}}^o_t, \mathrm{\textbf{r}}^d_t$ where $\mathrm{\textbf{r}}^o_t$ and $\mathrm{\textbf{r}}^d_t$ are ray's origin and direction respectively. At each time step, the intersections of stereo pairs of rays are calculated (green points), and they are used to guide the density of the rays during the optimisation. Then, samples along the rays (only from left frames) are taken by $\phi_{sampl}$, each sample is processed by $\phi_{c/\sigma}$ to predict the density $\sigma_t$ and colour $\mathrm{\textbf{c}}_t$ of each sample. A function $f_{ren}$ calculates the rendering weight from the density values as described in \cite{20_M} to render the colour of the pixels $\hat{\mathrm{\textbf{c}}}_t$. Similarly, the locations of the samples are merged into a single 3D point per pixel (blue points $\hat{\mathrm{{\textbf{P}}}}_{t_1}$ in the workspace at time step $t_1$ and yellow points $\hat{\mathrm{{\textbf{P}}}}_{t_2}$ in the workspace at time step $t_2$). The 3D points then go through a spatial transformation ($f_{cub}$) that contracts the workspace to a unit-cube. The invertible neural network ($\phi_{dist}$) performs the bidirectional mapping, considering both spatial and temporal information, between the workspace (deformable) and the canonical space (static). Thus, canonical points ($\mathrm{\textbf{P}}^c_{t_1}, \mathrm{\textbf{P}}^c_{t_2}$) are transformed to the time step opposite to the one from which they came, i.e. blue points in the workspace at $t_1$ are mapped to the canonical space then to $t_2$ ($\mathrm{\hat{\textbf{P}}}_{t_1 \rightarrow t_2}$), and yellow points in the workspace at $t_2$ are mapped to the canonical space then to $t_1$ ($\mathrm{\hat{\textbf{P}}}_{t_2 \rightarrow t_1}$). Finally, these points are expanded from the unit-cube and are projected to the image planes to obtain the 2D correspondences $\mathrm{\hat{\textbf{p}}}_{t_1 \rightarrow t_2}$ and $\mathrm{\hat{\textbf{p}}}_{t_2 \rightarrow t_1}$. Data flows in both directions ($t_1 \rightarrow t_2$ and $t_2 \rightarrow t_1$) during the optimisation to ensure consistency and chaining aggregated correspondences by minimising the correspondence error across all spaces (red lines in image planes, workspaces and canonical space).
However, after optimisation, $f_{LT}$ can be used, providing query points over the image plane at any time step $t_1$, a time step $t_2$, and camera poses for each $t$. 

The rest of this section first describes the data processing steps to obtain short-term correspondences, 3D information through triangulation and the tool's mask to simplify the optimisation. Here we focus on key differences compared to other TTO approaches.
Next, we describe the building blocks of the approach, which include a ray sampler, a density and colour predictor and an invertible network.
Finally, we discuss our optimisation approach, highlighting two aspects: a multi-scale approach for efficient pixel sampling and our loss function for cross-consistency across the image planes, deformable space and canonical space.

\subsection{Initial processing}
\subsubsection{Establishing short-term correspondences}
As described in \cite{17_M}, we obtain pixel correspondences between pairs of images using RAFT\cite{22_M} and classify each pixel correspondence as non-reliable, reliable, or occluded in a consistency mask $\mathrm{CM}_{t_1 \rightarrow t_2} \epsilon \{0,1,2\}$. However, we set a small maximum number of pairs per time step and leave a gap between the pairs of each time step to reduce memory usage. To address the limitation of RAFT, we use CoTrackerV3~\cite{78_M} to recover lost correspondences. We also experiment with MFT\cite{81_Wt}, which identifies the most reliable sequences (chains) of flows per pixel to enhance long-term correspondences, and allows us to use larger time-step gaps between image frames.

\subsubsection{Triangulation of target points}
Our approach leverages the camera's extrinsic parameters and reliable point correspondences between stereo images to guide the density of the rays projected from the camera models. So given pixel correspondences ($\textit{\textbf{p}}_{tL}, \textit{\textbf{p}}_{tR}$) between a stereo pair of images, the rays' origin and direction from those pixels are calculated to find the point of intersection. Since the rays of two matched pixels do not necessarily intersect, we define the intersection point as the closest point on the left rays to their matched right rays. In this way, the intersection can be defined by $\mathrm{\textbf{r}}^o_{tL} + \alpha \mathrm{\textbf{r}}^d_{tL} = \mathrm{\textbf{r}}^o_{tR} + \beta \mathrm{\textbf{r}}^d_{tR} + \gamma \hat{\mathrm{\textbf{d}}}_t$. Where $\hat{\mathrm{\textbf{d}}}_t$ is a unit vector along the shortest distance between the rays and the cross product between the ray's direction. This equation can be rewritten in its matrix form and solved as a system of three equations, so the intersection is $\mathrm{\textbf{P}}_t= \mathrm{\textbf{r}}^o_t + \alpha \mathrm{\textbf{r}}^d_t$. During training, the density of the ray is encouraged to be concentrated around this point.
Note that when the unit vector along the shortest distance between the rays is indeterminate, the rays are parallel. We mask out those pixels from the ray density optimisation function, allowing neighbouring pixels to define the density distribution in those regions.

\subsubsection{Workspace construction}
As described by Mip-NeRF 360~\cite{19_M}, we define a Axis-Aligned Bounding Box ($aabb$) for our use case that defines the workspace size, we compress the ray's samples into a unitary cube and shift their values to the positive coordinates of the Cartesian axis by using the function $f_{cub}(\mathrm{\textbf{r}}) = (\mathrm{\textbf{r}} - aabb_{min})/(aabb_{max}-aabb_{min})$ which is conveniently invertible. This allows our model to avoid abrupt responses to sign changes. 
Additionally, we encourage the canonical space to maintain the workspace boundaries by wrapping the unit cube with a unit sphere. In the optimisation section below, we detail how, during training, we bring back all the rays trying to escape from the limits of the unit sphere. This imposes the assumption that all the scenarios processed by our model strictly fit within the boundaries of a defined workspace, which is entirely valid for surgical scenarios.

\subsubsection{Tool masking}
To simplify the aggregation of information within InvNeRF, we identified the surgical instruments in every image of the video. Using SAM~\cite{79_M} we managed to segment the tools along each video using point prompts only on the first image of each video. We used this mask to segment out the tool during the scene optimisation and focus only on the tissues (deformable objects). 

\subsection{Model selection}
Ray sampling (Eq.~\ref{Eq_M:05}) is a crucial operation for fast rendering processing. In this work, we compare a fixed and evenly spaced sampling (which is not affected by deformations but requires a large number of samples) against a dynamic proposal network (DPN) to accelerate the process by taking fewer samples. DPN serves as the proposal network from NeRFAcc~\cite{20_M}, considering a 4th-dimensional input (after applying the space distortion function $f_{cub}$, we concatenate the normalised time steps).
\begin{equation}
\label{Eq_M:05}
\phi_{sampl}(\mathrm{\textbf{r}}^o_t, \mathrm{\textbf{r}}^d_t, t) = \mathrm{\textbf{r}}_t
\end{equation}

\begin{figure}[!t]
\centering
\includegraphics[width=3.4in]{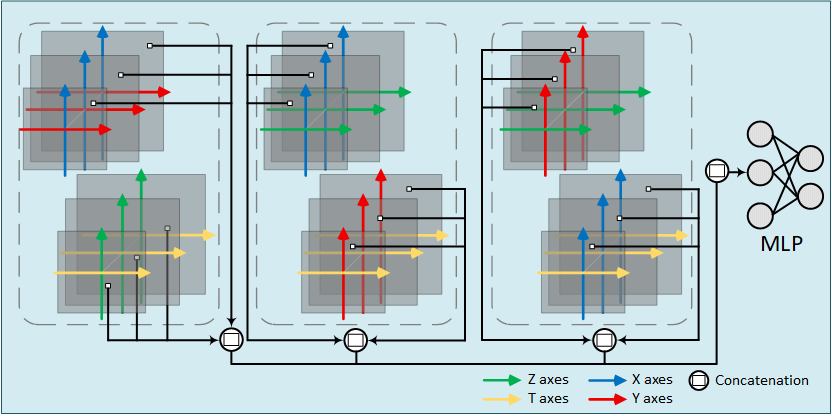}
\caption{\textbf{Multi-scale Hexa Plane}. The 4D coordinates of each pixel (x,y,z,t) are used to sample features from all planes. Features are concatenated and processed by an MLP.
}
\label{fig:4}
\end{figure}

For the colour and density estimator (Eq.~\ref{Eq_M:06}), we compared the performance of the network used in \cite{17_M}, a GaborNetwork, against our multi-scale HexPlanes (MHP) (Fig.~\ref{fig:4}). We built upon HexPlane~\cite{76_M}; we keep the structure, but we concatenate multiple resolutions with a small number of features. In this way, we reduced the number of parameters required to build the model and maintained the ability to capture fine details in density changes. We use two Multi-scale HexPlanes, one to encode the density values and one to encode the RGB values. Low-resolution planes quickly concentrate the density of all the rays to a distance where the object is likely to be located, and high-resolution planes take care of finer details. 

\begin{equation}
\label{Eq_M:06}
\phi_{c/d}(\mathrm{\textbf{r}}_t) = (\hat{\mathrm{\textbf{c}}}_t, \sigma_t)
\end{equation}

The invertible network (\ref{Eq_M:07}) has three main components: a feature generator (for spatial and temporal features), a network that processes those features and an invertible function that is parametrised by the network's response. We test the feature generators (encoders and feature planes) and invertible functions (CaDeX and CaDeX++) presented in \cite{17_M} and \cite{76_M}. However, we removed the use of homography transformations, reduced the number of layers, and used hyperbolic tangent activation functions in the network when using CaDeX.
\begin{equation}
\label{Eq_M:07}
\begin{array}{cc}
     \phi_{deform}(\hat{\mathrm{\textbf{P}}}_{t_1}, t_1) = \hat{\mathrm{\textbf{P}}}^c_{t_1} \\
    \phi_{deform}^{-1}(\hat{\mathrm{\textbf{P}}}^c_{t_1}, t_2) = \hat{\mathrm{\textbf{P}}}_{t_1 \rightarrow t_2} 
\end{array}
\end{equation}

\subsection{Optimisation}
This section describes our optimisation process for finding a function $f_{LT}$ parametrised by our invertible NeRF-based architecture, InNeRF. Along with the architecture that defines $f_{LT}$, our pixel sampling algorithm and proposed loss function are key components of our framework. The algorithm's objective is to reduce redundant pixel sampling, and our optimisation focuses on global motion estimation and geometrically consistent ray density. 

\subsubsection{Pixel sampling algorithm}
\begin{figure}[!t]
\centering
\includegraphics[width=3.5in]{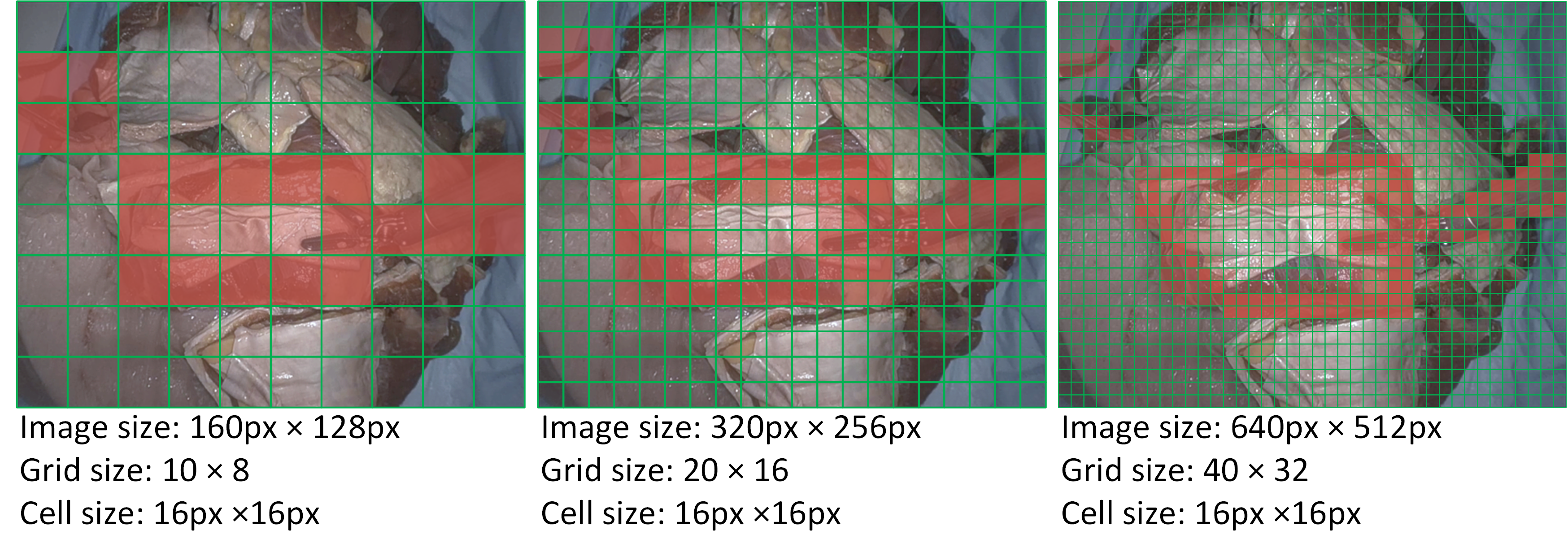}
\caption{\textbf{Multi-scale error grid sampling.}  Our error grid (green grid) begins at a coarse resolution, with pixel sampling concentrated in cells showing higher local error (red cells). As training progresses, the grid is refined to higher resolutions, focusing sampling on smaller, high-error regions. In the example, the images display the status of the normalised error grid immediately before the resolution is changed.
}
\label{fig:5}
\end{figure}

Our strategy to avoid redundant pixel sampling is to keep track of the 3D short-term correspondence error for each frame pair in a light error map. Algorithm~\ref{Alg:1} shows the steps to initialise the error map, to update it, and how to use it when sampling pixels at any iteration. The initialisation starts at the lowest requested scale, images are rescaled accordingly, and the dimensions of the rescaled images are used to initialise the error map ($e_{map}$) with a fixed cell size for each possible frame pair. The error map is updated every validation iteration, where for every frame pair, one random point within each cell ($e_{grid}$) is processed by $f_{LT}$, the error is calculated and used to update the error map. The reduced number of samples allows us to quickly perform this iteration and identify the areas where our function, $f_{LT}$, needs to focus. The average error over the map ($e_{current}$) is also used to decide when to upsample the size of the error map. The algorithm details how we sample pixels given a selected pair of frames with pixel correspondences based on the calculated error (Fig.~\ref{fig:5}). In addition, keeping track of the errors also allows us to consider cases where pairs of frames have already been solved. During optimisation, we select $t$ based on the error map related to that time step, thereby avoiding redundant frame sampling too.
It is important to notice that despite rescaling images and flow maps during the optimisation of the invertible NeRF representation, coordinates in the workspace remain consistent since the camera parameters are modified accordingly to the scales used during the optimisation.

\begin{algorithm}[H]
\caption{Pixel sampling algorithm}\label{Alg:1}
\begin{algorithmic}
\STATE {\textsc{Initialisation}}
\STATE \hspace{0.1cm} $n$: number of frames
\STATE \hspace{0.1cm} $m$: maximum number of pairs per frame
\STATE \hspace{0.1cm} $sz$: step-size, $s$: scale
\STATE \hspace{0.1cm} $H \leftarrow s \cdot H,\quad W \leftarrow s \cdot W$
\STATE \hspace{0.1cm}$ e_{map} \leftarrow   \textbf{0}~|~\textbf{0}  \in \mathds{R}^{n,2m,H/sz,W/sz} $
\STATE \hspace{0.1cm}$ e_{grid} \leftarrow  (i\cdot sz,j\cdot sz)~|~i \in [0,W/sz], j \in [0,H/sz]$
\STATE \hspace{0.1cm}$e_{total}\leftarrow inf$
\STATE {\textsc{Error map update (during validation)}}
\STATE \hspace{0.1cm}for $t \in [0,n]$
\STATE \hspace{0.3cm}for $k \in [- m,m]\backslash \{0\}$ do
\STATE \hspace{0.5cm}$p_{t} \leftarrow  e_{grid}+\Delta~|~ \Delta\sim  \mathcal{U}[0,sz]$
\STATE \hspace{0.5cm}$e \leftarrow  \mathcal{L}_{LT}(p_{t\rightarrow t+k},f_{LT}(p_{t},t,t+k))$
\STATE \hspace{0.5cm}$e_{map}(t,t+k) \leftarrow e$
\STATE \hspace{0.1cm}$e_{current}\leftarrow$ $\text{avg}(e_{map})$
\STATE \hspace{0.1cm}if $e_{current}>e_{total}$ and $s<1$ then:
\STATE \hspace{0.3cm} $H \leftarrow 2 \cdot H,\quad W \leftarrow 2 \cdot W$
\STATE \hspace{0.3cm} $e_{grid} \leftarrow (i \cdot sz, j \cdot sz)~|~i \in [0, W/sz], j \in [0, H/sz]$
\STATE \hspace{0.3cm} $e_{map} \leftarrow \text{interpolate}(e_{map},~\text{new shape}= H/sz \times W/sz)$
\STATE \hspace{0.3cm}  $s \leftarrow s\cdot 2$ 
\STATE \hspace{0.1cm}else:
\STATE \hspace{0.3cm} $e_{total}\leftarrow e_{current}$
\STATE {\textsc{Pixel sampling (Every optimisation iteration)}}
\STATE \hspace{0.1cm} $P_{t,t+k}(i,j)=\frac{e_{map}(t,t+k,i,j)}{\sum_{i,j} e_{map}(t,t+k,i,j)}$ 
\STATE \hspace{0.1cm} $\textbf{p}_t \leftarrow \left\{(i,j)~|~(i,j)\sim P_{t,t+k}\right\}$
\end{algorithmic}
\end{algorithm}

~ 

\subsubsection{Loss function}
We categorise the components of the loss function by the sub-spaces.

\textbf{Image plane}: We used an RGB loss $\mathcal{L}_{RGB}=|| \textbf{c}_{t_1}-\hat{\textbf{c}}_{t_1} ||^2 +|| \textbf{c}_{t_2}-\hat{\textbf{c}}_{t_2} ||^2$; here,  the rendered colour from our model is directly compared against the pixel colour on the images. On the image plane, we also evaluate the pixel correspondences between frames with the flow loss \cite{17_M} $\mathcal{L}_{flow}=|| \textit{\textbf{p}}_{t_1 \rightarrow t_2}-\hat{\textit{\textbf{p}}}_{t_1 \rightarrow t_2}||^2\mathds{1}_{\{\text{cm}_{t_1 \rightarrow t_2} \in [1,2]\}} + || \textit{\textbf{p}}_{t_2 \rightarrow t_1}-\hat{\textit{\textbf{p}}}_{t_2 \rightarrow t_1}||^2\mathds{1}_{\{\text{cm}_{t_1 \rightarrow t_2} \in [1]\}}$. This function masks out non-reliable correspondences using the consistency mask when moving from $t_1$ to $t_2$. In the opposite direction, the occluded regions are also masked out because the correspondences are calculated from $t_1$ to $t_2$, so the predicted points $\hat{\textit{\textbf{p}}}_{t_2 \rightarrow t_1}$ from the non-occluded pixels in $I_2$ do not have a correspondence. 
\begin{equation}
\label{Eq:05}
\mathcal{L}_{2D} =  \lambda_{flow}\mathcal{L}_{flow} + \lambda_{RGB}\mathcal{L}_{RGB}
\end{equation}

\textbf{Workspace}: 
Inspired by EndoNeRF~\cite{06_Wd}, and given the limited motion of the camera in a surgical scenario, we incorporate a Gaussian target loss to guide the density of the rays around triangulated 3D target points from stereo images. Given the target points $\mathrm{\textbf{P}}_{t_1}$, rays origin and samples along the rays, we estimate an ideal density around the triangulated depth points, $\sigma_{target}=e^{-(||\mathrm{\textbf{P}}_{t_1}-\mathrm{\textbf{r}}_{t_1}^o||-||\mathrm{\textbf{r}}_{t_1}-\mathrm{\textbf{r}}_{t_1}^o||)/2\sigma_{std}}$ and a target rendering weights as $w_{target} = \sigma_{target}/||\sigma_{target}||$. Finally, we minimise the error between the target weights and the rendering weight predicted from the predicted density on the ray's samples using a Kullback-Leibler divergence loss, $\mathcal{L}_{GT} = w_{target}(log(w_{target})-log(\hat{w}))$.


Similarly to Omnimotion, we also add a smooth flow loss, but in the tracked 3D points, resulting in a smooth displacement along X, Y, and Z, $\mathcal{L}_{smooth}=||(\hat{\mathrm{\textbf{P}}}_{t_1\rightarrow t_1+1} - \hat{\mathrm{\textbf{P}}}_{t_1}) - (\hat{\mathrm{\textbf{P}}}_{t_1} - \hat{\mathrm{\textbf{P}}}_{t_1\rightarrow t_1-1})||$.

\begin{equation}
\label{Eq:06}
\mathcal{L}_{3D} = \lambda_{GT}\mathcal{L}_{GT} + \lambda_{smooth}\mathcal{L}_{smooth} 
\end{equation} 

\textbf{Canonical space}: The sphere loss constraint the canonical object to the size of the deformed workpace. Every canonical sample that is pushed away from this volume is then dragged back to its non-deformed state, $\mathcal{L}_{CR} = || \mathrm{\textbf{P}}^c_{t_1}-\mathrm{\textbf{r}}_{t_1} ||^2\mathds{1}_{||2\mathrm{\textbf{P}}^c_{t_1}-1|| > 1}$.

As a complementary loss to the geometric consistency we minimise the distance between correspondences in the canonical space since both should represent the same location in this space, $\mathcal{L}_{XC} =||\mathrm{\textbf{P}}^c_{t_1}-\mathrm{\textbf{P}}^c_{t_2} ||^2 \mathds{1}_{\{\text{cm}_{t_1\rightarrow t_2} \in [1]\}}$.

\begin{equation}
\label{Eq:11}
\mathcal{L}_{canon} =  \lambda_{CR}\mathcal{L}_{CR} + \lambda_{XC}\mathcal{L}_{XC}
\end{equation}

\section{Experiments}

\subsection{Datasets}
The STIR (Surgical Tattoos in Infrared) dataset~\cite{01_Dt} is designed for evaluating tissue tracking algorithms using infrared (IR) markers created with indocyanine green (ICG) dye. Videos are captured with a da Vinci Xi surgical robot’s stereo laparoscope in both visible and IR modes, showing tissue before and after manipulation. 
The validation dataset was released as part of the STIR Challenge 2024~\cite{80_Dt}. It includes 60 video clips, recorded at 25 fps, with durations ranging from 0 to 60 seconds, and clean 2D and 3D labels for point tracking. 

The SCARED (Stereo Correspondence and Reconstruction of Endoscopic Data) dataset~\cite{69_Dd} consists of endoscopic videos recorded at 25 frames per second (fps) from fresh porcine cadaver abdomens using a da Vinci Xi system, paired with depth maps obtained via structured light projection. Although the scenes are static (with no tissue deformation or instrument movement), the camera moves during the videos, and the ground-truth camera poses are provided for each pair of stereo images. This data is used to validate the integration of the camera pose in our framework and the estimation of 3D points. 
\subsection{Evaluation metrics}
We report the average end-point error ($L^2$-norm) between the tracked points and the ground truth across videos. The 2D tracking error is reported in pixels (using the original image resolution), and the 3D tracking error is reported in millimetres ($mm$). The average training time is also reported. For the comparison vs feed forward methods, we also report the accuracy percentage per threshold ($\delta_n$): $\delta_{4px}$, $\delta_{8px}$, $\delta_{16px}$, $\delta_{32px}$ and $\delta_{64px}$ on the image plane and  $\delta_{2mm}$, $\delta_{4mm}$, $\delta_{8mm}$, $\delta_{16mm}$, $\delta_{32mm}$ on the workspace. Finally, the structural similarity index (SSIM) and the Peak Signal-to-Noise Ratio (PSNR) are reported for the rendered images. 

\subsection{Implementation details} 
The experiments presented in this paper were conducted using NVIDIA L40S 48GB GPUs provided by the HPC service, Aire, at the University of Leeds. We used a batch size of 16 frame pairs, 256 pixels per sample, 32 samples per projected ray, a cell size of 16$\times$16 pixels in the error map, a learning rate of $1\times10^{-4}$, and Adam optimiser. An epoch is completed when the model has seen all the images in the video, and we update the error map every 25 epochs. The proposed framework comprises two schedulers and one early-stop criterion, which track the average and std of the error map. The first scheduler upscales the resolution of the error map on a plateau by a factor of $2$ (as described in algorithm~\ref{Alg:1}), using a patience of 2 epochs, an improvement tolerance of $0.01$, and a starting scale factor of 0.25. The second scheduler was used for updating the learning rate on a plateau, with three epochs of patience, a factor of $0.5$, and an improvement tolerance of $0.01$. Finally, the stopping criterion is activated once the scale factor of the error map is 1, using a patience of 6 epochs and an improvement tolerance of $0.001$. 

\subsection{Results}
\begin{table*}[!t]
\caption{\textbf{Test-time optimisation SOTA comparison.}}
\centering
\begin{tabular}{llllllllll}
\hline
\\[-1em]
Model            & $\phi_{deform}$& $\phi_{c/d}$ & $\phi_{sampl}$ & ptsB(2D) & ptsB(3D) & ptsA(3D) & train-time & PSNR & SSIM\\
                 &                &             &                 & \scriptsize{avg {${l}^2${-norm}}} & \scriptsize{avg {${l}^2${-norm}}} & \scriptsize{avg {${l}^2${-norm}}} & avg &          & \\ \hline
Omnimotion~\cite{17_M}& MLP-CaDex      & GaborNet    & \multicolumn{1}{l|}{Fixed}   & 31.793$_{px}$      & ---         & ---         & 22.01$_{hrs}$        & ---         & --- \\
FastOmniTrack~\cite{77_M}    & Plane-CaDex++  & Pred Depth  & \multicolumn{1}{l|}{Fixed}   & 26.863$_{px}$      & ---         & ---         & 2.00$_{hrs}$          & ---         & ---\\ \hline
\multicolumn{10}{l}{Our approach}                                                                                                                 \\ \hline
InvNeRF           & MLP-CaDex      & GaborNet    & \multicolumn{1}{l|}{Fixed}   & 17.103$_{px}$      & ~7.929$_{mm}$      & ~3.831$_{mm}$                     & 2.38$_{hrs}$           & 21.006 & 0.509 \\
InvNeRF           & MLP-CaDex      & HexaPlane   & \multicolumn{1}{l|}{Fixed}   & 17.534$_{px}$      & ~6.979$_{mm}$      & ~3.764$_{mm}$                     & 2.13$_{hrs}$           &\textbf{22.496} & \textbf{0.556} \\
InvNeRF           & MLP-CaDex      & HexaPlane   & \multicolumn{1}{l|}{DPN-32}  & 19.909$_{px}$      & 19.639$_{mm}$      & 19.109$_{mm}$                     & 1.80$_{hrs}$           & 17.553 & 0.440 \\
InvNeRF           & MLP-CaDex      & HexaPlane   & \multicolumn{1}{l|}{DPN-16}  & 22.350$_{px}$      & 46.475$_{mm}$      & 44.822$_{mm}$                     & \textbf{1.42$_{hrs}$}   & 17.576 & 0.441 \\
InvNeRF           & MLP-CaDex++    & HexaPlane   & \multicolumn{1}{l|}{Fixed}   & 16.919$_{px}$      & ~6.776$_{mm}$      & ~3.537$_{mm}$                     & 2.21$_{hrs}$        & 22.489 & \textbf{0.556} \\
InvNeRF           & Planes-CaDex++ & HexaPlane   & \multicolumn{1}{l|}{Fixed}   & 18.255$_{px}$      & ~7.099$_{mm}$      & ~3.648$_{mm}$                     & 1.80$_{hrs}$           & 22.479 & \textbf{0.556} \\
InvNeRF$_\text{mask}$  & MLP-CaDex++    & HexaPlane   & \multicolumn{1}{l|}{Fixed}   & \textbf{13.925$_{px}$} & \textbf{~6.202$_{mm}$} & \textbf{~3.205$_{mm}$} & 2.03$_{hrs}$& 22.430 & \textbf{0.557} \\ \hline   
\end{tabular}
\label{table:1}
\end{table*}

A thorough evaluation was done to assess the different aspects of our approach. First, we present the results of our pixel sampler approach. Then, a comparison of our framework with the state-of-the-art test-time optimisation approaches for 2D tracking is presented. Due to the limitations of Omnimotion~\cite{17_M}'s training speed and memory usage, we perform this comparison over a small subset of the original STIR dataset (15 videos with fewer than 175 frames) and a temporal window of 16 consecutive frames as defined by FastOmniTrack~\cite{77_M}. Here, we also present how the inclusion of some of the most important ideas in neural rendering (discussed in the previous section) affects our framework's performance. Using the validation dataset from STIR Challenge 2024~\cite{80_Dt}, we present a comparison, including FastOmniTrack and other feed-forward tracking methods presented in the challenge.
Unlike the mentioned tracking models, our model can render and track the scene at full resolution and receive the extrinsic camera parameters from a robotic system, so we further evaluated the accuracy of our model using the scarce 3D ground truth provided in the training SCARED dataset.

 \begin{figure}[h!]
\centering
\includegraphics[width=3.5in]{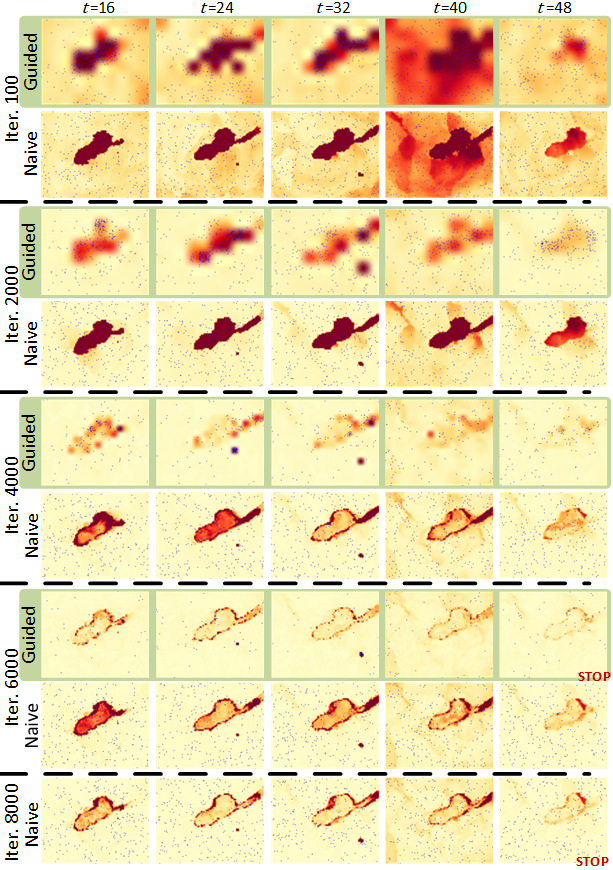}
\caption{\textbf{Error map comparison during optimisation} between our guided and a naive approach. Deeper reds signify higher short-term 2D pixel correspondence error. Our guided approach converges with low error at 6000 iterations, while the naive (used by Omnimotion and FastOmniTrack) approach is stopped after 8000 iterations without converging.
}
\label{fig:6}
\end{figure}

Fig.~\ref{fig:6} shows the optimisation comparison of the model over a video when using the 3D short-term correspondence error map for sampling pixels (Guided) against using the random pixel sampler from other TTO methods (Naive). In the shown example, the error-map sampling converges (plateaus) significantly earlier and stops. In contrast, the other approach continues training for longer without reaching the status of the error map in our approach (the read areas in iteration 8000 are visibly more intense than in our approach at iteration 6000) and stops when reaching the maximum number of iterations. In this way, our approach reduces the optimisation time and makes better use of the precomputed correspondences. 

\begin{figure}[t]
\centering
\includegraphics[width=3.5in]{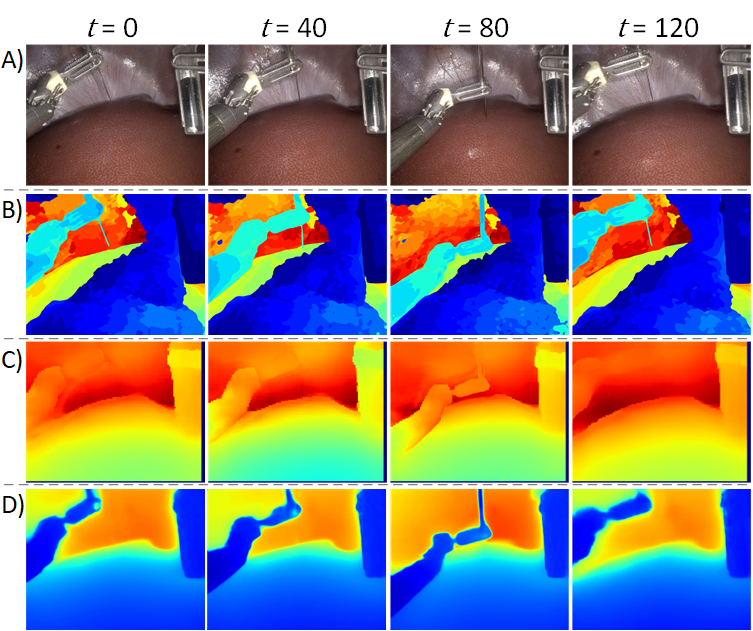}
\caption{\textbf{Qualitative depth comparisons of test-time optimisation SOTA.} RGB images (A) and depth maps (B,C,D) at times $t = $, 0, 40, 80, and 120 $s$ from a sequence. OmniMotion~\cite{17_M} (B) and FastOmniTrack~\cite{77_M} (C) assume an orthographic camera and do not offer real-world scaling. Our approach, InvNeRF (D), uses a pinhole camera and offers a metric-scale depth map.
}
\label{fig:7}
\end{figure}

Our main comparison is against other state-of-the-art rendering techniques for point tracking. Table~\ref{table:1} presents the error of these approaches and our framework, with different configurations (discussed in the previous sections). 
Column ptsB(2D) presents the tracking error over the image plane, whereas column ptsB(3D) and ptsA(3D) present the 3D errors for 3D point tracking and 3D rendered location, respectively. Our experiments demonstrate the superiority of our Multi-scale HexPlanes for rendering 3D location in a deformable scenario. They also showed that despite the fact that a proposal network reduces the number of samples in the ray and speeds up the optimisation process, it struggles to accurately concentrate the density of the rays in the areas of interest. Finally, they also showed that the use of a non-linear function (CaDex++) without extra data structures is the best to model the deformations in a surgical scenario. Overall, the combination of MLP-CaDex++, our MHP and a fixed sampling in our proposed framework (InvNeRF) showed the best performance. Masking the tools out of the optimisation (InvNeRF$_{(mask)}$) further improves the results of our approach by a 42\% in 2D tracking; nevertheless, our approach in all configurations surpasses the TTO SOTA and is the first TTO approach for 3D tracking too. The results in 3D show an average location rendering error of 3.2$mm$ and tracking error of 6.2$mm$, bringing us closer to achieving a consistent and accurate 3D tracking of deformable surgical scenarios. Our approach recovers the benefits of using render-based approaches for image rendering, so we also report the metrics to assess reconstructed images. We reconstruct the right image of the stereo cameras every 10 time steps and compare it against the original right images (these images were not used during the optimisation).   
Fig.~\ref{fig:7} shows the difference in the reconstructed depth of the three models; Omnimotion's does not hold any relationship with the depth to the actual surface of the objects, which is probably one of the main reasons it takes so long for the model to find a solution. FastOmniTrack directly utilises precomputed scale-agnostic depth over an orthographic camera model and tracks it, which is more consistent with the object's surface; however, it is unreliable for 3D reconstruction. Our approach has been demonstrated to be an alternative that utilises a pinhole camera model and a NeRF-based model to aggregate 3D data from a dynamic camera and generate accurate and metric estimation of the object's location.

\begin{figure}[h]
\centering
\includegraphics[width=3.4in]{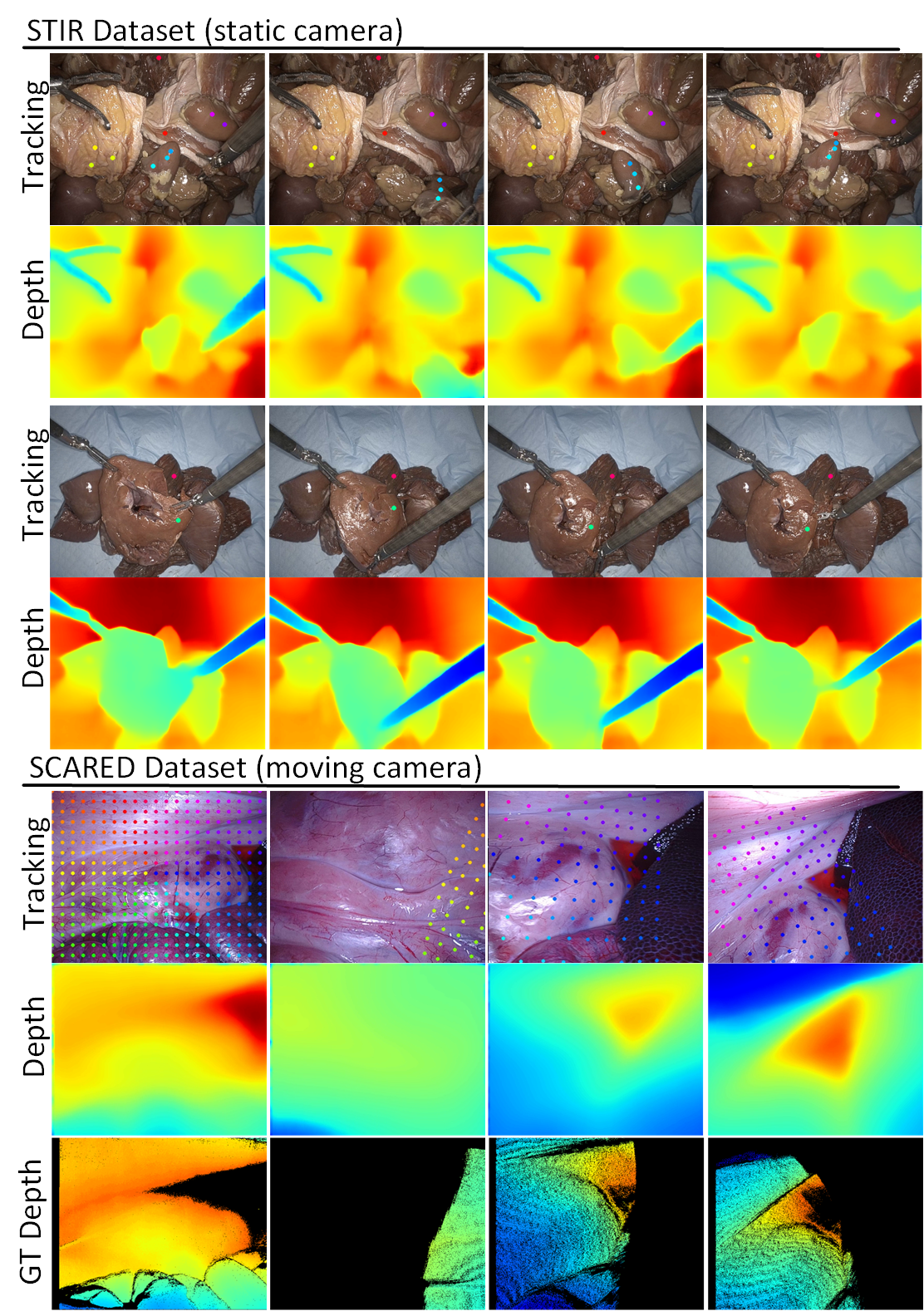}
\caption{\textbf{Qualitative results over the STIR and SCARED datasets.} 
}
\label{fig:8}
\end{figure}

On the comparison using the full validation set presented in STIR Challenge 2024~\cite{80_Dt}, we included FastOmniTrack, the feed forward approaches that competed in the challenge, our top performance configuration (InvNeRF with fixed sampler, MHP and MLP-CaDex++), this configuration with masked tools (InvNeRF$_\text{mask}$) and with MTF correspondences using a time-step gap of 8 frames (InvNeRF$_\text{MFT}$), Table~\ref{table:2}. Among the render-based methods, our approach maintains the top performance. With the use of MFT correspondences, we further improve the results, which demonstrate that the TTO approaches will always benefit from every improvement made to obtain reliable correspondences. The accuracy difference compared to MFT may be due to the inclusion of sub-optimal chains in our approach, which MFT removes in their per-pixel flow analysis. We can also see that, despite FastOmniTrack's good performance for the lowest threshold, it falls short in the others, which explains the general lower average when compared to our approach and demonstrates higher consistency in ours, regardless of the complexity of the deformations the tissues undergo. For 3D tracking (Table~\ref{table:3}), we only include the different configurations of our framework since FastOmnitrack is not metric. Compared to other methods that focus solely on point tracking, our method surpasses other approaches while retaining the advantages of rendering-based methods, including the inference of optical flow in both 2D and 3D, as well as 3D reconstruction and image synthesis. 
Fig.~\ref{fig:8} show some qualitative tracking and depth results on the STIR datasets of challenging scenes where the tissues are occluded (first tracking and depth rows), move fast (second tracking and depth row), so the correspondences drift. We can see that the tracked points are not lost, the drift is small, and the depth information is consistent with the movements of the tissue. 

\begin{table}[h]
\caption{Comparison of InvNeRF against non-rendering based methods on 2D point tracking}
\begin{tabular}{lllllll}
\hline
\\[-1em]
\multicolumn{1}{l|}{Method}  & $\delta^4_{px}$  & $\delta^8_{px}$  & $\delta^{16}_{px}$ & $\delta^{32}_{px}$ & $\delta^{64}_{px}$ & $\delta_{avg}$ \\ \hline
\multicolumn{6}{l}{Baseline}                                                                                                            \\ \hline
\multicolumn{1}{l|}{RAFT}    & 7.26            & 19.56           & 39.92           & 64.92           & 80.44           & 42.42          \\
\multicolumn{1}{l|}{CSRT}    & 22.78           & 47.38           & 67.14           & 74.80           & 81.05           & 58.63          \\
\multicolumn{1}{l|}{MFT}     & 42.54           & 69.36           & 86.49           & \textbf{93.35}  & \textbf{96.37 } & 77.62          \\ \hline
\multicolumn{6}{l}{Miccai 2024 competitors}                                                                                             \\ \hline
\multicolumn{1}{l|}{ICVS\_2AI}     & 25.40          & 51.01          & 74.19          & 88.71 & 92.54           & 66.37        \\
\multicolumn{1}{l|}{JMEES}         & 26.00          & 54.44          & 77.42          & 91.43 & 95.36           & 68.99        \\
\multicolumn{1}{l|}{MFTIQ$^{1}$}   & 42.34          & 68.95          & 85.89          & 92.14 & 94.76           & 76.82        \\
\multicolumn{1}{l|}{MFTIQ$^{2}$}   & \textbf{44.36} & \textbf{69.56} & 85.89          & 91.13 & 95.16           & 77.22        \\ 
\multicolumn{1}{l|}{CUHK}          & 40.93          & 68.95          & \textbf{87.50} & 93.15 & \textbf{96.37 } & 77.38        \\\hline
\multicolumn{6}{l}{Test-time optimisation approaches}                                                                                  \\\hline
\multicolumn{1}{l|}{FastOmniTrack}      & 31.17 & 51.04 & 71.54 & 83.47 & 90.16 & 65.48\\
\multicolumn{1}{l|}{InvNeRF}            & 27.02 & 47.81 & 73.38 & 86.90 & 93.65 & 65.69\\
\multicolumn{1}{l|}{InvNeRF$_\text{mask}$} & 25.91 & 51.60 & 77.94 & 89.99 & 94.86 & 68.05\\ 
\multicolumn{1}{l|}{InvNeRF$_\text{MFT}$}  & 34.90 & 61.96 & 84.56 & 91.49 & 96.64 & 73.39\\ \hline
\end{tabular}
\label{table:2}
\end{table}

\begin{table}[h]
\caption{Comparison of InvNeRF against non-rendering-based methods on 3D point tracking}
\begin{tabular}{lllllll}
\hline
\\[-1em]
\multicolumn{1}{l|}{Method}            & $\delta^2_{mm}$ & $\delta^4_{mm}$ & $\delta^8_{mm}$ & $\delta^{16}_{mm}$ & $\delta^{32}_{mm}$ & $\delta_{avg}$  \\ \hline
\multicolumn{6}{l}{Baseline}                                                                                                                    \\ \hline
\multicolumn{1}{l|}{RAFT-Stereo}       & 13.94          & 36.16          & 60.40          & 79.80           & 91.51           & 56.36           \\ \hline
\multicolumn{6}{l}{Miccai 2024 competitors}                                                                                                     \\ \hline
\multicolumn{1}{l|}{JMEES}             & 25.70          & 45.40          & 65.31          & 81.16           & 91.01           & 61.71           \\
\multicolumn{1}{l|}{ICVS\_2AI}         & 27.88          & 55.15          & 75.96          & 91.11           & 97.58           & 69.54           \\ \hline
\multicolumn{6}{l}{Test-time optimisation approaches}                                                                                                   \\ \hline
\multicolumn{1}{l|}{InvNeRF}           & 22.03          & 49.27          & 74.01          & 88.98           & \textbf{99.16}  & 66.69           \\ 
\multicolumn{1}{l|}{InvNeRF$_\text{mask}$}& 27.19          & 53.74          & \textbf{76.87} & 91.00           & 98.92           & 69.55  \\
\multicolumn{1}{l|}{InvNeRF$_\text{MFT}$} & \textbf{32.66} & \textbf{57.49}          & 76.51          & \textbf{92.39}  & 98.65           & \textbf{71.54}           \\ \hline
\end{tabular}
\label{table:3}
\end{table}

Since the STIR dataset lacks camera position information, and all experiments had to be run with a static camera, we also tested our model on the videos of the SCARED dataset. Fig.~\ref{fig:8} shows the comparison between the ground truth depth and the rendered depth. We evaluate the accuracy of the depth over the pixels with a valid ground truth, having an average 3.05 $mm$ error, and more than 83\% of the pixels with an error below 5 $mm$. These results are comparable to those of recent supervised methods on single frames and videos from the same dataset. The second part of Fig.~\ref{fig:8} shows the qualitative results for tracking and depth over one sequence. We can see that depth is consistent with the portion of ground truth depth provided in the dataset and that this section also matches the tracked tissue on the 2D images.

\begin{table}[h]
\caption{Ablation study}
\setlength{\tabcolsep}{0.5em} 
\begin{tabular}{lll|lllll}
\hline
\\[-1em]
GTL & CCL & SFL & PSNR            & ptsB(2D)       & ptsB(3D)        & ptsA(3D)       & Time           \\
    &     &     &                 & \scriptsize{avg $l^2$-norm}         & \scriptsize{avg {${l}^2${-norm}}}          & \scriptsize{avg $l^2${-norm}}         & avg            \\ \hline
\\[-1em]
             & \checkmark   & \checkmark   & 20.319          & 32.201         & 47.538         & 50.441         & \textbf{0.855} \\
\checkmark   &              &              & \textbf{22.453} & 18.170         & 9.736          & 3.252          & 1.957          \\
\checkmark   & \checkmark   &              & 22.446          & 14.421         & 6.582          & \textbf{3.168} & 1.958          \\
\checkmark   &              & \checkmark   & 22.430          & 18.011         & 8.312          & 3.222          & 1.815          \\
\checkmark   & \checkmark   & \checkmark   & 22.429          & \textbf{13.925}& \textbf{6.202} & 3.206          & 2.035 \\ \hline
\end{tabular}
\label{table:4}
\end{table}

Finally, the ablation study is presented in Table~\ref{table:4}, which evaluates the influence of the different components in our loss function.

\section{Discussion and conclusion}
We present the first NeRF-based test-time optimisation to address 2D and 3D tracking simultaneously and the first algorithm to avoid redundant pixel sampling on TTO, making the optimisation of surgical scenes more efficient. In the surgical context, our method outperforms any other 2D tracking TTO method, reducing the mean average 2D error by 50\% (Table~\ref{table:1}), and it does not fall far behind other feed-forward 2D tracking approaches (Table~\ref{table:2}). It also outperforms all methods on 3D tracking by 5\% on the average end-point-error (Table~\ref{table:3}) and retains the advantages of rendering-based techniques. Thus, optical flow in both 2D and 3D, as well as 3D reconstruction of dynamic and deformable scenarios, is encoded in the implicit representation of our InvNeRF (Figure~\ref{fig:7}). A key distinction of our method from other NeRF-based approaches is that we only need to render each new visualised location once, after which we track it. This advantage removes many of the artefacts produced in other approaches. 

Our framework maintains a consistent relation with the deformable space by constraining the workspace size and linking 3D correspondences to a small volume in the canonical space. However, the topology of the objects and the geodesic distances in the canonical space could be improved in future works. This essentially could improve the convergence and ability to track points under extreme complex deformations or prolonged periods of occlusion. 
Finally, as for other TTO methods, any improvement on 2D correspondences will also further improve the representations created by our InvNeRF method.

In conclusion, our approach establishes the state of the art with large improvement in 3D point-tracking compared to existing techniques in surgical scenarios. This is the first method that encodes representations from structure, colour and 3D flow of the object making it applicable towards estimation of accurate points tracking and related depth estimation. We demonstrate that our results are comparable to those of methods that specialise in only one of these tasks.
\bibliographystyle{IEEEtran}
\bibliography{references}
\end{document}